\documentclass[sigconf]{acmart}

\usepackage{multirow}
\usepackage{booktabs}
\usepackage{graphicx}
\usepackage{enumitem}
\AtBeginDocument{%
  \providecommand\BibTeX{{%
    \normalfont B\kern-0.5em{\scshape i\kern-0.25em b}\kern-0.8em\TeX}}}

\copyrightyear{2022}
\acmYear{2022}
\setcopyright{acmcopyright}
\acmConference[ICMR '22]{Proceedings of the 2022 International Conference on Multimedia Retrieval}{June 27--30, 2022}{Newark, NJ, USA}
\acmBooktitle{Proceedings of the 2022 International Conference on Multimedia Retrieval (ICMR '22), June 27--30, 2022, Newark, NJ, USA}
\acmPrice{15.00}
\acmDOI{10.1145/3512527.3531359}
\acmISBN{978-1-4503-9238-9/22/06}

\begin{document}
\title{3D-Augmented Contrastive Knowledge Distillation for Image-based Object Pose Estimation}

\author{Zhidan Liu}
\orcid{0000-0002-8958-3139}
\email{zdliu20@fudan.edu.cn}
\affiliation{%
  \institution{School of Computer Science, Fudan University}
  \city{Shanghai}
  \country{China}
}
\author{Zhen Xing}
\orcid{0000-0001-6407-0321}
\email{zxing20@fudan.edu.cn}
\affiliation{%
  \institution{School of Computer Science, Fudan University}
  \city{Shanghai}
  \country{China}
}
\author{Xiangdong Zhou}
\email{xdzhou@fudan.edu.cn}
\affiliation{%
  \institution{School of Computer Science, Fudan University}
  \city{Shanghai}
  \country{China}
}
\author{Yijiang Chen}
\orcid{0000-0001-9727-8305}
\email{chenyj20@fudan.edu.cn}
\affiliation{%
  \institution{School of Computer Science, Fudan University}
  \city{Shanghai}
  \country{China}
}
\author{Guichun Zhou}
\email{19110240014@fudan.edu.cn}
\affiliation{%
  \institution{School of Computer Science, Fudan University}
  \city{Shanghai}
  \country{China}
}

\renewcommand{\shortauthors}{Zhidan Liu, et al.}
\fancyhead{}

\begin{abstract}

Image-based object pose estimation sounds amazing because in real applications the shape of object is oftentimes not available or not easy to take like photos. Although it is an advantage to some extent, un-explored shape information in 3D vision learning problem looks like ``flaws in jade''. In this paper, we deal with the problem in a reasonable new setting, namely 3D shape is exploited in the training process, and the testing is still purely image-based. We enhance the performance of image-based methods for category-agnostic object pose estimation by exploiting 3D knowledge learned by a multi-modal method. Specifically, we propose a novel contrastive knowledge distillation framework that effectively transfers 3D-augmented image representation from a multi-modal model to an image-based model. We integrate contrastive learning into the two-stage training procedure of knowledge distillation, which formulates an advanced solution to combine these two approaches for cross-modal tasks. We experimentally report state-of-the-art results compared with existing category-agnostic image-based methods by a large margin (up to +5\% improvement on ObjectNet3D dataset), demonstrating the effectiveness of our method.

\end{abstract}




\maketitle

\begin{figure}[hpt]
\includegraphics[width=\columnwidth]{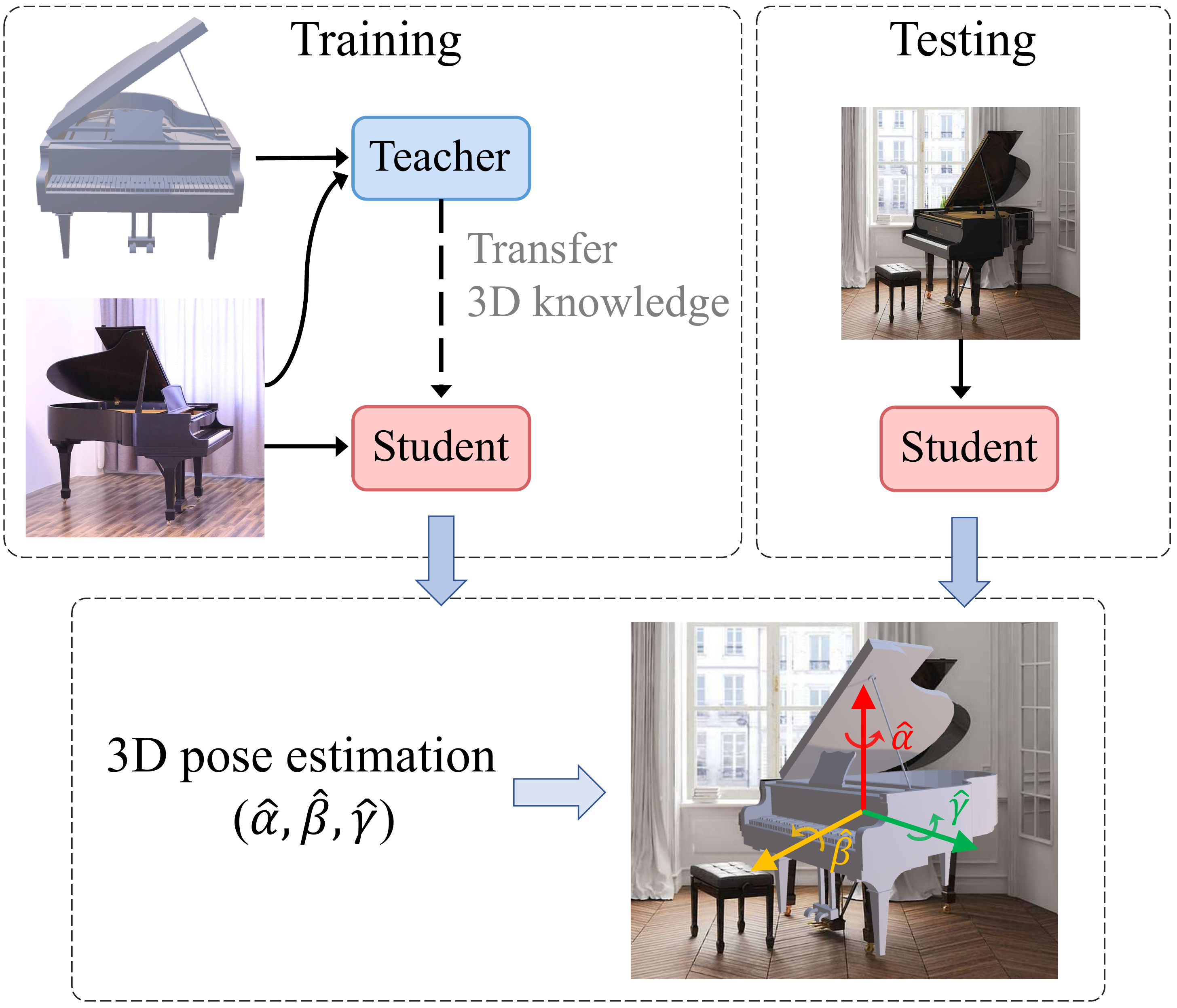}
\centering
\caption{Abstract illustration of our proposed method. We first train a teacher model on 3D shapes and images, then we transfer its 3D knowledge to a student model trained solely on images. Thereby, employing only the image-based student model at testing, we enhance the performance on images. For category-agnostic object pose estimation, the output consists of three angles, representing the viewpoint of the pictured object regardless of its category.}
\label{fig:illustration}
\end{figure}

\section{INTRODUCTION}
Rigid object pose (viewpoint) estimation shows great potential in many exciting practical applications such as robotic manipulation \cite{collet2011moped, zhu2014single, tremblay2018deep}, augmented reality \cite{marchand2015pose, tang20193d, Fu_2021} and autonomous driving \cite{wu20196d, manhardt2019roi}, etc. The main stream on this topic is image-based category-agnostic method \cite{grabner20183d, zhou2018starmap, Xiao2019PoseFS, pitteri2019cornet, Xiao2020PoseContrast}, 
because it is more generalized when encountering objects of novel categories in applications. In the traditional object pose estimation, 3D shapes are usually exploited as additional input to leverage the 3D geometry information to improve the estimation performance \cite{Xiao2019PoseFS, park2020latentfusion, dani20213dposelite}. However, 3D shape is usually unavailable and acquiring 3D shapes is time consuming and labor intensive on-site. Therefore, the pure image-based object pose estimation without any 3D shape information has emerged \cite{grabner20183d, pitteri2019cornet, Xiao2020PoseContrast, zhou2018starmap, icmrpose_1, icmrpose_2}. In these works, one common approach \cite{grabner20183d, zhou2018starmap, icmrpose_1, icmrpose_2} is to leverage keypoint features for pose estimation after detecting and re-projection of 2D keypoints, which requires for a suitable design of category-agnostic keypoints on various object geometries. Another approach \cite{Xiao2020PoseContrast} applies contrastive learning \cite{chen2020simple} to exploiting RGB-based geometric similarities shared between various categories, which heavily depends on the feature representation. However, it is noted that these methods are far inferior to those works of exploiting 3D shape information \cite{Xiao2019PoseFS, park2020latentfusion, dani20213dposelite}.

We note that with the progress of 3D vision learning, more and more shape data will be available when we prepare the training sets off-site \cite{xiang2016objectnet3d, xiang2014beyond, sun2018pix3d}. Therefore, in this paper we deal with the object pose estimation problem in a novel and reasonable setting, that is 3D shape can be exploited in the training process, and the testing is still purely image-based. It means we can enhance the performance of an image-based model by utilizing 3D information in the training process. Figure \ref{fig:illustration} gives more illustrations on our motivation. In our work 3D shapes are explored by a teacher model and the learned 3D knowledge is transferred to an image-based student model. Hence, employing only the image-based model at inference, the implicit 3D shape information can be utilized to enhance the performance of the pose estimation. Apparently to accomplish this, we need to deal with the challenge of cross-modal knowledge distillation. There are some previous works studied on related problems \cite{lopez2016unifying, gupta2016cross}, and most of the previous works adopt the teacher-student model for knowledge distillation. However, to our best knowledge, few works deal with the cross-modal distillation on object pose estimation with the context of category-agnostic. It is more challenging than the traditional knowledge distilling problem.

In this work we present a teacher-student model with a novel contrastive learning based bridge layer to deal with pose estimation in a cross-modal setting. Different from traditional knowledge distillation framework, we manage to alleviate the difficulty of transferring knowledge from different domains (3D shape and image) by using an intermediate latent space. Instead of directly transferring knowledge from the 3D shape augmented space, we first apply contrastive learning to build a ``bridge'' latent space for the first step of dealing with the cross-modal; second we employ the teach-student distilling mechanism to transfer the knowledge from the bridge space to the image-based student model. The success of contrastive learning \cite{chen2020simple, tian2019contrastive} in self-supervised learning tasks invokes a great deal of research interests. It has been applied to cross-modal representation learning that allows the interplay across features in different modalities \cite{alayrac2020self, patrick2021compositions, radford2021learning, yuan2021multimodal, zhang2020counterfactual, kim2021learning}. Therefore we adopt contrastive learning as the bridge layer. 

Specifically, in our framework we train the student to capture 3D-augmented structural knowledge in the teacher's representation of the data. However, knowledge distillation methods require a two-stage training procedure for the teacher and the student respectively, while end-to-end contrastive learning methods \cite{chen2020simple, wu2018unsupervised, oord2018representation} train a two-tower network simultaneously.  To deal with the problem, as shown in Figure \ref{fig:architecture}, we first combine the pose estimation task and the contrastive learning as a joint objective, encouraging the Contrastive Learner (bridge layer module) to obtain the learned 3D information. Afterwards, by freezing the pre-trained teacher model including the bridge layer module, we transfer its knowledge to the student model via knowledge distillation.

In summary, we make the following contributions.
\begin{itemize}[leftmargin=*]
\item We present a novel cross-modal teacher-student knowledge distillation framework to deal with object pose estimation problem in a cross-modal setting. That is the learned 3D shape information is transferred to an image-based student model to boost the performance of the category-agnostic image-based object pose estimation.
\item We formulate an advanced solution to alleviate the gap between different data modalities by effectively combining contrastive learning and knowledge distillation. Specifically, we propose the Contrastive Learner as the ``bridge'' across two models to best distill privileged cross-modal knowledge, solving the contradiction of training procedures between two approaches.
\item We experimentally show state-of-the-art results compared with existing category-agnostic image-based methods in different settings by a large margin (up to +5\% improvement on ObjectNet3D dataset), demonstrating the effectiveness of our method.
\end{itemize}

\section{RELATED WORK}

\noindent\textbf{Category-agnostic Object Pose Estimation.} 3D object pose estimation is a challenging problem in computer vision and robotics \cite{kendall2015posenet, pavlakos20176}, etc. Significant progress is achieved in the deep learning era. The category-specific methods are first proposed \cite{su2015render, tulsiani2015viewpoints, mousavian20173d, kundu20183d, grabner2019gp2c, wang2019normalized}. In these methods, an independent prediction branch is built for each object category. However, it suffers from the problem of limited labeled object categories. To overcome the novel object categories problem, a few category-agnostic methods \cite{grabner20183d, zhou2018starmap, Xiao2019PoseFS, pitteri2019cornet, Xiao2020PoseContrast} have emerged in recent years. In contrast to category-specific methods, these methods focus on exploiting the shared features regardless of the category information. Therefore, without knowing specific category of the object, category-agnostic methods are relatively generalized when testing on novel categories. 

The category-agnostic methods can be further classified into two categories: one is grounded on making use of both 3D shapes and images, the assumption is that the 3D shapes are available in both of training and testing \cite{Xiao2019PoseFS, park2020latentfusion, dani20213dposelite}. For leveraging 3D shapes, \cite{Xiao2019PoseFS} aggregates 3D shape and RGB image information for arbitrary objects, representing 3D shapes as multi-view renderings or point clouds. Similarly, \cite{dani20213dposelite} proposes a lighter version of \cite{Xiao2019PoseFS} by encoding the 3D shapes into graphs using node embeddings \cite{grover2016node2vec}. However, these multi-modal methods are limited as 3D shapes are oftentimes unavailable at testing \cite{xiang2016objectnet3d, xiang2014beyond, sun2018pix3d}; The other category is the image-based methods \cite{grabner20183d, pitteri2019cornet, Xiao2019PoseFS, Xiao2020PoseContrast, zhou2018starmap, icmrpose_1, icmrpose_2}, that is only images are exploited for pose estimation. \cite{grabner20183d, pitteri2019cornet} regard corners of the 3D bounding box as generic keypoints, which only focus on cubic objects with simple geometric shape. \cite{zhou2018starmap} obtains the rotation by weighting keypoint distance by the heatmap value. These keypoint-based methods are robust but fail with heavy occlusions and tiny or textureless objects. In order to exploit 2D geometry features shared between various categories, \cite{Xiao2019PoseFS} proposes a simple but effective coarse-to-fine method for predicting viewpoints based on ResNet \cite{lin2017feature}, and \cite{deng2020self, mustikovela2020self, wang2020self6d} conduct self-supervised learning on unlabeled images. Inspired by \cite{Xiao2019PoseFS, deng2020self, mustikovela2020self, wang2020self6d}, \cite{Xiao2020PoseContrast} utilizes contrastive learning for pose estimation with MOCOv2 \cite{chen2020improved}. However, these methods exhibit inferior performance to those harnessing 3D shapes.

\noindent\textbf{Generalized Knowledge Distillation.} The purpose of knowledge distillation \cite{hinton2015distilling} is to transfer knowledge from one model (teacher) to another model (student) \cite{heo2019comprehensive, zhang2019your, park2019relational, tian2019contrastive}. In basic methods, a supplemental supervision is imposed by asking the student to minimize the Kullback-Leibler (KL) divergence between its output prediction distribution and the teacher's.

The concept is further extended to the generalized distillation \cite{lopez2016unifying} for learning with privileged information \cite{vapnik2009new} together with knowledge distillation. In generalized knowledge distillation, one uses distillation to extract useful knowledge from the privileged information of the teacher \cite{finn2017model}. \cite{lopez2016unifying, gupta2016cross} initially propose the technique of transferring knowledge between images from different modalities. In our work, the 3D shape information is only available at the training, but the obtained 3D shape knowledge can be transferred for the testing by using generalized knowledge distillation.

\noindent\textbf{Contrastive Learning.} By maximizing mutual information between related signals in contrast to others, contrastive learning methods \cite{wu2018unsupervised, oord2018representation, tian2020contrastive, chen2020simple, chen2020big} are able to learn powerful image features. Among the various forms of the contrastive loss functions \cite{wang2015unsupervised, hadsell2006dimensionality, hjelm2018learning, wu2018unsupervised, oord2018representation}, InfoNCE \cite{oord2018representation} has become a common pick in many methods. These contrastive learning methods make advanced results of unsupervised learning on ImageNet \cite{deng2009imagenet}. More recent works employ contrastive learning for multi-modal inputs \cite{alayrac2020self, patrick2021compositions, radford2021learning, yuan2021multimodal, zhang2020counterfactual, kim2021learning} and demonstrate its effectiveness in cross-modal representation learning. \cite{kim2021learning} proposes a cross-modal contrastive learning framework for domain adaptation, treating each modality as a view in contrastive learning. Inspired by \cite{kim2021learning}, our approach treats 3D modality as a view, leveraging a contrastive learning objective for performing feature regularization mutually among different feature spaces. 

\noindent\textbf{Knowledge Distillation meets Contrastive Learning.} Recently, some works \cite{tian2019contrastive, sun2020contrastive, dai2021learning, tejankar2021isd} combine contrastive learning and knowledge distillation. \cite{tian2019contrastive} introduces a contrastive loss to transfer structural knowledge of the teacher network, providing a new perspective for better distilling knowledge from intermediate representations. However, they conduct contrastive learning after freezing the teacher model, leading to one-side learning of the student model. It is limited when model structures differ widely between two models. In our method, we propose to indirectly transfer knowledge learned by joint contrastive learning, which formulates an advanced combination of these two approaches.

\section{METHODOLOGY}

In this section, we present the proposed \emph{3D-augmented contrastive knowledge distillation for image-based object pose estimation} (3DAug-Pose). We start by introducing multi-modal model architectures for the pose estimation task. Then we propose to transfer knowledge from the teacher to the image-based student model by means of knowledge distillation and contrastive learning.

\subsection{Feature Extraction and Fusion}
\begin{figure*}[htp]
\includegraphics[width=17cm]{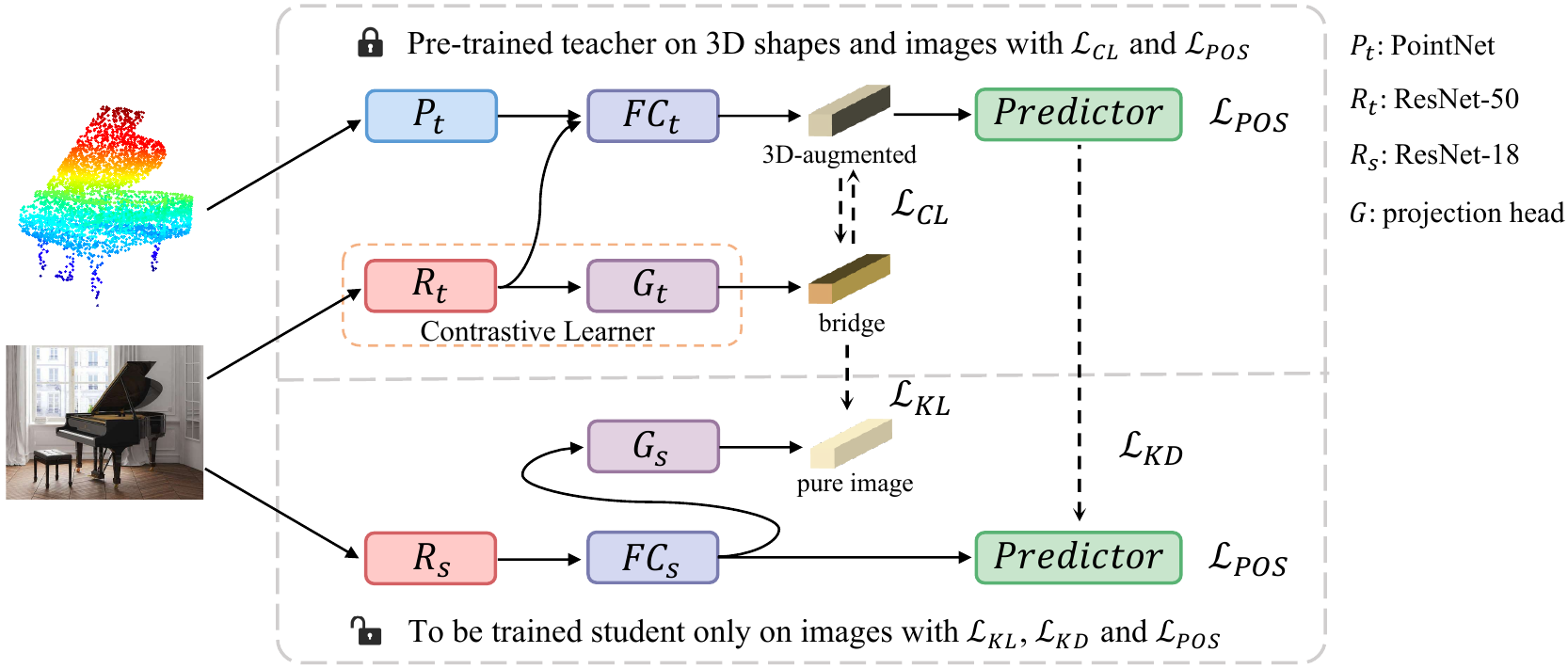}
\centering
\caption{Overview of our contrastive knowledge distillation framework. Given RGB images and corresponding 3D shapes (point cloud), we train the teacher model with object pose estimation loss $\mathcal{L}_{POS}$ and cross-modal contrastive loss $\mathcal{L}_{CL}$. Afterwards, we train the student model solely on RGB images guided by the frozen teacher model (including the Contrastive Learner) with $\mathcal{L}_{KD},\mathcal{L}_{KL}$ and $\mathcal{L}_{POS}$, obtaining a 3D-augmented image representation space that is optimized for object pose estimation. Note that the Contrastive Learner conducts contrastive learning with $\mathcal{L}_{CL}$, generating ``bridge representation'' learned from 3D-augmented image representation, which is further transferred to the image-based student model with $\mathcal{L}_{KL}$.}
\label{fig:architecture}
\end{figure*}

In terms of different input modalities, we adopt two models with different structures, namely teacher and student. As shown in Figure \ref{fig:architecture}, the teacher model consists of a 3D encoder $P_t$ and an image encoder $R_t$, while the student model only relies on an image encoder $R_s$. Thereafter, the teacher model is able to capture 3D-augmented information by effectively fusing representations of two modalities.

\noindent\textbf{3D and Image Encoders.} For the student model, as the input excludes 3D shapes, we only utilize an image encoder $R_s(\cdot)$ (ResNet-18 \cite{He_2016_CVPR}) for extracting image features from images $\mathcal{X}$. Represented as $x_s$, the image feature vectors are later passed to fully connected layers $FC_s(\cdot)$ with each layer (2048-800-400-200) followed by Batch Normalization and ReLU activation:
\begin{equation}
x_s = R_s(\mathcal{X}),
\end{equation}
\begin{equation}
h_s = FC_s(x_s).
\end{equation}

For the teacher model, given images $\mathcal{X}$ of objects paired with corresponding 3D shapes $\mathcal{D}$ (point cloud), we make use of two separate feature extractors: $R_t(\cdot)$ (ResNet-50) to encode $\mathcal{X}$ and $P_t(\cdot)$ (PointNet \cite{Qi_2017_CVPR}) to encode $\mathcal{D}$. Therefore, representations of two modalities are obtained as $x_t$ and $d_t$ by the teacher model:
\begin{equation}
x_t = R_t(\mathcal{X}),
\end{equation}
\begin{equation}
d_t = P_t(\mathcal{D}).
\end{equation}

\noindent\textbf{Multi-Modal FuseNet.} As aforementioned, the teacher model obtains representations consisting of image feature vectors $x_t$ and 3D feature vectors $d_t$. In order to aggregate 3D and image information to enhance representation, we aim at fusing features of both modalities. Similar to \cite{Xiao2020PoseContrast, Xiao2019PoseFS, pan2019deep}, we adopt fully connected layers $FC_t(\cdot)$ with non-linear activation ReLU on the first three layers and \emph{tanh} on the final output layer. Note that it also downsizes the dimension of aggregated features for the further classification:
\begin{equation}
h_t = FC_t(d_t\  concat \  x_t).
\end{equation}

\subsection{Pose Estimation} 
\label{pose_estimation}

3D rotation matrix $R$ of the pictured object is decomposed into three Euler angles as in \cite{su2015render, Xiao2019PoseFS}: azimuth $\alpha$, elevation $\beta$ and in-plane rotation $\gamma$, with $ \alpha,\gamma \in [-\pi,\pi)$ and $\beta \in [-\pi/2, \pi/2]$. Similar to recent works \cite{Xiao2019PoseFS, xiao2020few, Xiao2020PoseContrast}, we consider the prediction task as a coarse-to-fine classification problem. Specifically, we split each Euler angle $\theta \in {\{\alpha,  \beta, \gamma\}}$ uniformly into discrete bins $i$ of bin size $B$ (=$\pi/2$ in our experiments). The model outputs include bin classification scores $b_{\theta,i}  \in [0,1]$ and offset proportions $ \delta_{\theta,i}  \in [0,1]$ within each bin.

Therefore, we feed the downsized feature vector $h_t$ or $h_s$ into multiple predictors consisting of linear layers. Based on the outputs, we adopt a cross-entropy loss for angle bin classification and a smooth-L1 loss for bin offset regression:
\begin{equation}
\mathcal{L}_{POS} =  \sum_{ \theta  \in {\alpha,  \beta, \gamma}} \mathcal{L}_{cls}({\rm bin}_\theta,b_\theta)+ \mathcal{L}_{reg}({\rm offset}_\theta, \delta_\theta),
\end{equation}
where ${\rm bin}_\theta$ is the ground-truth bin and ${\rm offset}_\theta$ is the offset proportion for angle $\theta$. Then the final prediction for angle $\theta$ is:
\begin{equation}
\hat{\theta} = (j +  \delta _{\theta, j})B \quad {\rm with} \quad j={\rm arg\mathop{max}\limits_{i}} b_{\theta,i},
\end{equation}
where $i \in [-12,...,11]$ for $\alpha,\gamma$, and $i \in [-6,...,5]$ for $\beta$.

\begin{figure}[t]
\includegraphics[width=8cm]{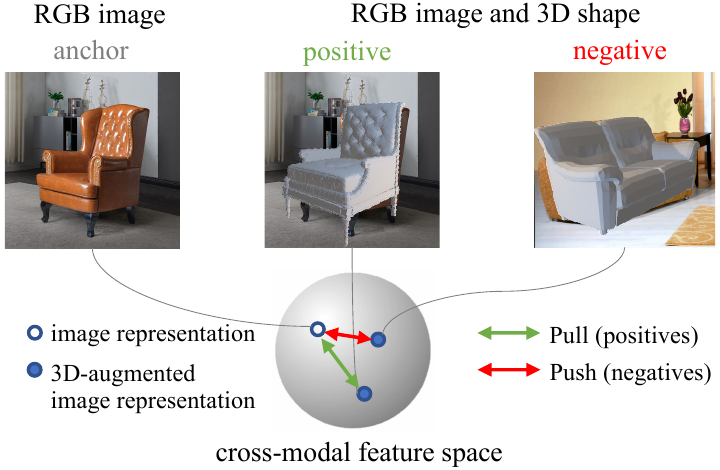}
\centering
\caption{An example of cross-modal contrastive learning. After encoding and projecting multi-modal inputs to the shared cross-modal feature space, we pull together the RGB image (e.g., lounge chair) and its 3D-augmented variant as positives while pushing it apart from negatives (different 3D-augmented images, e.g., sofa).}
\label{fig:contrastive}
\end{figure}

\subsection{Contrastive Knowledge Distillation}
\label{contrastive_knowledge_distillation}

In this subsection, we aim at enhancing the performance of the image-based student model on the object pose estimation task via knowledge distillation. To this end, we propose a novel method that can distill 3D knowledge learned through contrastive learning, namely \emph{contrastive knowledge distillation}. Employing the implicit 3D information, the student model can perform better on image data at testing. 

Since this requires good initial pose estimates and different input modalities, we rely on a two-stage training procedure. As shown in Figure \ref{fig:architecture}, we start by training the teacher model taking advantage of 3D shapes, thus it achieves excellent performance by establishing coherence between 3D shapes and RGB images. Afterwards, we freeze the teacher model and transfer its knowledge to the student model trained solely on RGB images. Note that we consider cross-modal knowledge distillation from a two-level perspective. Besides classification output guidance of the teacher model, we propose to combine contrastive learning methods for better distilling knowledge from intermediate layer representations. Moreover, we yield pose-related data augmentation to further enhance the efficiency of knowledge distillation.

\noindent\textbf{Contrastive Learner.} In order to better distill implicit 3D information, we focus on transferring knowledge in the latent feature space. Inspired by \cite{kim2021learning, yuan2021multimodal}, we leverage contrastive learning methods for representation learning of cross-modal data.

However, combining contrastive learning and knowledge distillation suffers from a severe gap of training procedures. While most knowledge distillation methods rely on the two-stage training procedure, end-to-end contrastive learning methods train a two-tower network simultaneously. Nonetheless, previous works \cite{tian2019contrastive, sun2020contrastive, dai2021learning} ignore this difference and 
directly conduct contrastive learning at the second training stage. Although they preserve representation ability for downstream tasks by freezing the teacher model, the performance of contrastive learning is limited due to the one-side learning of the student. Especially when the student model structure differs widely from the teacher's, we need to consider about involving the teacher model during contrastive learning.

Therefore, we propose a Contrastive Learner $C(\cdot)$ as a bridge layer module for joint contrastive learning and indirect knowledge transferring. Specifically, by regarding the fused feature vectors $h_t$ as 3D-augmented image feature vectors, we attempt to let $C(\cdot)$ learn from $h_t$ jointly and later transfer them to the student model. To this end, we adopt the teacher's image encoder $R_t(\cdot)$ as the main form of $C(\cdot)$. As shown in Figure \ref{fig:architecture}, since $C(\cdot)$ is grounded on the teacher model, we can conduct cross-modal contrastive learning to adjust $P_t(\cdot)$ and $R_t(\cdot)$ simultaneously. In this way, we suppose the Contrastive Learner obtains ``bridge representation'' across two models. Afterwards, we freeze the pre-trained $C(\cdot)$ and transfer the learned ``bridge representation'' to the student's image encoder $R_s(\cdot)$. Note that $C(\cdot)$ and $R_s(\cdot)$ have similar structures (both based on ResNet), thus it can yield great distillation efficiency through this bridge layer module.

We consider a batch of $N$ samples consisting of $\{\mathcal{X}_i\}_{i=1}^N$ and $\{\mathcal{D}_i\}_{i=1}^N$, where $\mathcal{X}_i$ and $\mathcal{D}_i$ represent the RGB image and the 3D shape of the $i$th object sample, respectively. In the context of multi-modal encoders $R_t(\cdot)$, $P_t(\cdot)$ and the FuseNet $FC_t(\cdot)$, the teacher model generates image feature vectors $\{x_{ti}\}_{i=1}^N$ and 3D-augmented image feature vectors $\{h_{ti}\}_{i=1}^N$. As shown in Figure \ref{fig:contrastive}, for cross-modal representation learning, the key here is to use contrastive loss function to encourage the representations learned for the same training sample to be similar. In other words, the contrastive loss learns to minimize the difference between $x_{ti}$ and $h_{ti}$. In the mean time, for different sample $i$ and $j$, the contrastive loss maximizes the difference between $x_{ti}$ and $h_{tj}$. To this end, we first leverage $C(\cdot)$ consisting of $R_t(\cdot)$ and a projection head $G_t(\cdot)$ \cite{chen2020simple} that maps representations to the space where contrastive loss is applied. We use fully connected layers with two hidden layers as $G_t(\cdot)$ to obtain the projected $x_{ti}$ as $z_{ti}$:
\begin{equation}
z_{ti} = G_t(x_{ti}),
\end{equation}
where $z_{ti}$ has the same dimension as $h_{ti}$. Then we treat $(z_{ti},h_{ti})$ as positive pairs and $(z_{ti},h_{tj})$ as negative pairs for $i\, \neq\, j$. Let $s(z_{ti}, h_{ti}) =  \langle z_{ti}, h_{ti} \rangle / (||z_{ti}|| \cdot ||h_{ti}||)$. To encourage the above properties, we define the contrastive loss for a batch of $N$ samples as:
\begin{equation}
\mathcal{L}_{CL} = -\frac{1}{N}  \sum_{i \in [N]} {\rm log}\, \frac{{\rm exp}\,(s(z_{ti},h_{ti})/ \tau )}{ \sum\nolimits_{j \in [N]} {\rm exp}\,(s(z_{ti},h_{tj})/ \tau )},
\end{equation}
where $\tau$ is a temperature parameter \cite{chen2020simple}.

\noindent\textbf{Intermediate Representation Distillation.} As $C(\cdot)$ effectively captures the ``bridge representation'' from 3D-augmented image representation by contrastive learning at the first stage, our aim is to transfer its knowledge to the student's image encoder $R_s(\cdot)$ at the second stage. We speculate that the deepest layers closer to the bottleneck, which are related to higher-level properties for the object pose estimation task, should be more similar. Based on that, we propose to impose the privileged 3D knowledge of $C(\cdot)$ to the student model in its bottleneck by making their latent representations as close as possible. Specifically, for the student's compressed image feature vectors $h_s$, we pass them through a projection head $G_s(\cdot)$ to obtain the projected $h_s$ as $z_s$:
\begin{equation}
z_s = G_s(h_s),
\end{equation}
where $G_s(\cdot)$ consists of fully connected layers with one hidden layer followed by ReLU activation, and these $z_s$ have the same dimension as $h_s$ and $z_t$.

Given a batch of $N$ image samples $\{\mathcal{X}_i\}_{i=1}^N$, we apply the Kullback-Leibler (KL) divergence as a loss function between $z_t$ and $z_s$:
\begin{equation}
\mathcal{L}_{KL} = D_{KL}(z_t||z_s) =  \sum_{i \in [N]} \sum_{j \in [E]} z_{ti}^j \, {\rm log}  \frac{z_{ti}^j}{z_{si}^j},
\end{equation}
where $z_{ti},z_{si}$ are previously normalized, and $E$ denotes the embedding dimension of $z_{ti}$ and $z_{si}$. Note that this function is not symmetric and in this order it is to make $z_{si}$ similar to $z_{ti}$ generated by the frozen $C(\cdot)$ based on the same image sample $\mathcal{X}_i$.

\noindent\textbf{Output Distribution Distillation.} From the perspective of distilling knowledge in output distribution, the term $\mathcal{L}_{KD}$ is added to the loss function. Here, we adopt KL divergence to measure the correspondence between two networks' predictions. Assuming classification predictions of the frozen teacher and the student network are denoted as $p_{t}$ and $p_{s}$ respectively, the loss term is given as:
\begin{equation}
\mathcal{L}_{KD} = D_{KL}(p_t||p_s) =  \sum_{i \in [N]} \sum_{c \in [C]} p_{ti}^c \, {\rm log}  \frac{p_{ti}^c}{p_{si}^c},
\end{equation}
where $N$ is the batch size, and $C$ is the number of prediction classes (angle bins). $p_{i}^c$ refers to the predicted probability of the $c$th class for the $i$th sample.

\noindent\textbf{Pose-Related Data Augmentation.} In order to further enhance the efficiency of knowledge distillation and improve the performance of object pose estimation, we yield pose-related data augmentation consisting of rotation and horizontal flip as in \cite{Xiao2020PoseContrast}. Specifically, an image rotation of angle $\varphi$ refers to in-plane rotation angle $\gamma+\varphi$ for the object, and a horizontal image flip refers to a change of sign of azimuth $\alpha$ and in-plane rotation $\gamma$. By increasing the quantity of training data for the second training stage, we consider it as a proper way to better perform our approaches.

\subsection{Training Loss}

Overall, we detail loss functions in subsection \ref{pose_estimation} and \ref{contrastive_knowledge_distillation} for the object pose estimation task and contrastive knowledge distillation approaches. Consequently, the combined loss function for training the teacher model is:
\begin{equation}
\mathcal{L}_{teacher} = \kappa_1 \mathcal{L}_{POS} +  \kappa_2 \mathcal{L}_{CL},
\end{equation}
where $\kappa_1,\kappa_2$ denote balance factors for $\mathcal{L}_{POS}$ and $\mathcal{L}_{CL}$, respectively.

Then, in the second training stage, the combined loss function for training the student model is:
\begin{equation}
\mathcal{L}_{student} = \omega_1 \mathcal{L}_{POS} +   \omega_2 \mathcal{L}_{KL} + \omega_3 \mathcal{L}_{KD},
\end{equation}
where $\omega_1,\omega_2,\omega_3$ denote balance factors for $\mathcal{L}_{POS}$, $\mathcal{L}_{KL}$ and $\mathcal{L}_{KD}$, respectively. Our model is robust to these parameters. We refer to subsection \ref{experimental_setup} for more details on the hyper-parameters.

\begin{table*}[t]
\caption{Experimental results of category-agnostic object pose estimation on ObjectNet3D and Pascal3D+. The experiments are conducted in the fully-supervised setting with all categories seen. \cite{Xiao2019PoseFS} proposes multiple approaches, in which PoseFromShape${{}^{\textbf{+}}}$ takes 3D shapes as additional input. *StarMap actually obtains the rotation by solving for a similarity transformation between the image frame and the world frame, weighting keypoint distances by the heatmap value.}
\begin{tabular}{lccccccc}
\toprule[1pt]
\multirow{2}{*}{Method}  & \multirow{2}{*}{Test w/ 3D} & \multirow{2}{*}{PnP} & \multirow{2}{*}{Backbone} & \multicolumn{2}{c}{ObjectNet3D} & \multicolumn{2}{c}{Pascal3D+} \\ \cmidrule(r){5-6} \cmidrule(r){7-8}
                         &                             &                      &                           & Acc30 ↑        & MedErr ↓       & Acc30 ↑       & MedErr ↓      \\ \midrule[1pt]
3DPoseLite \cite{dani20213dposelite}               & \checkmark                           &                      & ResNet-18                 & -              & -              & 0.80          & 13.4          \\ 
PoseFromShape${{}^{\textbf{+}}}$ \cite{Xiao2019PoseFS}      & \checkmark                           &                      & ResNet-18                 & 0.74           & 18.1           & 0.82          & 10.8          \\ \specialrule{0em}{0.5pt}{0pt} \hline \specialrule{0em}{0.5pt}{0pt} 
Grabner et al. \cite{grabner20183d}          &                             & \checkmark                    & ResNet-50                 & -              & -              & 0.81          & 11.5          \\ 
PoseContrast \cite{Xiao2020PoseContrast}         &                             &                      & ResNet-50                 & 0.66           & 28.8           & 0.81          & 11.9          \\ 
StarMap \cite{zhou2018starmap}                 &                             & \checkmark*                   & ResNet-18                 & 0.56           & 42.1           & \textbf{0.82} & 12.8          \\ 
PoseFromShape \cite{Xiao2019PoseFS} &                             &                      & ResNet-18                 & 0.65           & 31.6           & 0.79          & 12.6          \\ 
3DAug-Pose (Ours)                &                             &                      & ResNet-18                 & \textbf{0.70}  & \textbf{25.6}  & \textbf{0.82} & \textbf{10.9} \\ \bottomrule[1pt]
\end{tabular}
\label{fully-supervised}
\end{table*}

\begin{table*}[t]
\caption{Cross-dataset evaluation of category-agnostic object pose estimation on Pix3D. The methods are trained on Pascal3D+ training set and tested on Pix3D, where 6 categories are unseen (novel) and 3 categories are already seen.}
\begin{tabular}{llccccccccccc}
\toprule[1pt]
\multirow{2}{*}{}  &  \multirow{2}{*}{Method}         & \multirow{2}{*}{Test w/ 3D} & \multicolumn{6}{c}{NOVEL}                                                                                            & \multicolumn{3}{c}{SEEN}                                                         & \multirow{2}{*}{Mean} \\ \cmidrule(r){4-9} \cmidrule(r){10-12}
\multicolumn{2}{l}{}                                &                             & tool          & misc          & b-case        & \multicolumn{1}{l}{bed} & desk          & \multicolumn{1}{l}{w-drobe} & \multicolumn{1}{l}{table} & \multicolumn{1}{l}{sofa} & \multicolumn{1}{l}{chair} &                       \\ \midrule[1pt]
\multirow{5}{*}{\rotatebox[origin=c]{90}{Acc30 ↑}} & 3DPoseLite \cite{dani20213dposelite}              & \checkmark                          & 0.09          & 0.10          & 0.62          & 0.58                    & 0.66          & 0.57                        & 0.40                      & 0.94                     & 0.50                      & 0.50                  \\ 
                         & PoseFromShape${{}^{\textbf{+}}}$ \cite{Xiao2019PoseFS}       & \checkmark                           & 0.07          & 0.28          & 0.71          & 0.54                    & 0.71          & 0.65                        & 0.53                      & 0.94                     & 0.79                      & 0.58                  \\ \specialrule{0em}{0.5pt}{0pt} \cline{2-13} \specialrule{0em}{0.5pt}{0pt} 
                         & PoseFromShape \cite{Xiao2019PoseFS}  &                             & 0.04          & 0.15          & 0.72          & 0.65                    & 0.73          & 0.51                        & 0.52                      & 0.92                     & 0.79                      & 0.56                  \\ 
                         & PoseContrast \cite{Xiao2020PoseContrast}            &                             & 0.04          & 0.18          & 0.62          & 0.60                    & \textbf{0.76} & \textbf{0.54}               & \textbf{0.53}             & 0.93                     & 0.78                      & 0.55                  \\
                         & 3DAug-Pose (Ours)               &                             & \textbf{0.07} & \textbf{0.19} & \textbf{0.82} & \textbf{0.66}           & 0.75          & 0.52                        & 0.52                      & \textbf{0.94}            & \textbf{0.80}             & \textbf{0.59}         \\ \bottomrule[1pt]
\end{tabular}
\label{cross-dataset}
\end{table*}

\section{EXPERIMENT}

\subsection{Experimental Setup}
\label{experimental_setup}
\noindent\textbf{Datasets.} We conduct experiments on three challenging and commonly used datasets for benchmarking object pose estimation in the wild. Although they all feature various objects and environments, they differ largely in the quality of pose annotations and aligned 3D shapes. ObjectNet3D \cite{xiang2016objectnet3d} contains 100 categories in a subset of ImageNet \cite{deng2009imagenet} with relatively accurate pose annotations and delicate 3D shapes aligned for images. Therefore, we consider ObjectNet3D as our major evaluation benchmark. Pascal3D+ \cite{xiang2014beyond} contains only 12 rigid categories of PASCAL VOC 2012 \cite{everingham2010pascal} with approximate pose annotations due to coarsely aligned 3D shapes. Pix3D \cite{sun2018pix3d} proposes a small dataset with only 9 categories. It includes image-shape pairs with pixel-level 2D-3D alignment, improving the quality of annotations. Following \cite{tulsiani2015viewpoints,su2015render,Xiao2019PoseFS,Xiao2020PoseContrast}, we test our methods only on non-occluded and non-truncated objects.

\noindent\textbf{Implementation Details.} All our experiments are implemented using PyTorch. As for hyper-parameters, we set the dimension of object image features and shape features to 1024 for the multi-modal teacher model and set the dimension of object image features to 2048 for the image-based student model. We use parameters $\tau=0.1$, $\kappa_1=1$, $\kappa_2=0.5$, $\omega_1=0.25$, $\omega_2=0.75$, $\omega_3=0.75$ and the transformation rotation $\varphi$ varies in [-15°,15°]. We train all our networks using the Adam optimizer with initial learning rate of 1e-4, which is divided by 10 at 80\% of the training phase. In the first stage, we train the teacher model with batch size as 160 for 300 epochs. In the second stage, we freeze the teacher model and train the student model with batch size as 46 for 90 epochs.

\noindent\textbf{Evaluation Metrics.} Following \cite{tulsiani2015viewpoints, su2015render}, we compute common metrics: Acc30 is the percentage of estimations with rotation error less than 30 degrees; MedErr is the median angular error in degrees.

\subsection{Experimental Results}
\label{experimental_results}
In this subsection, we present the experimental results in different settings: fully-supervised, cross-dataset, zero-shot and few-shot, demonstrating the superiority and generalizability of our approach. 

\noindent\textbf{Fully-Supervised Setting.} We first conduct experiments in the fully-supervised setting, where all the evaluation categories are seen during training. We consider it as a way to let the teacher model take the most advantage of existing 3D shapes by using the training set and then boost the performance of the student model.

We first conduct the experiments on two datasets: ObjectNet3D \cite{xiang2016objectnet3d} and Pascal3D+ \cite{xiang2014beyond}. Following the common protocol \cite{grabner20183d,zhou2018starmap}, we train our models on the training set and test them on the val set, with both sets sharing the same categories. As shown in Table \ref{fully-supervised}, we compare with state-of-the-art methods of category-agnostic object pose estimation. Note that we list both multi-modal and image-based methods. For multi-modal methods that utilize 3D shapes as additional input, PoseFromShape${{}^{\textbf{+}}}$ \cite{Xiao2019PoseFS} adopts 3D point cloud as our teacher model does, showing the effectiveness of employing 3D shapes. For image-based methods, PoseContrast \cite{Xiao2020PoseContrast} originally uses the MOCOv2 pre-trained backbone \cite{chen2020improved}, but here we replace it with a randomly initialized one for a fair comparison.

\begin{figure*}[htp]
\includegraphics[width=16cm]{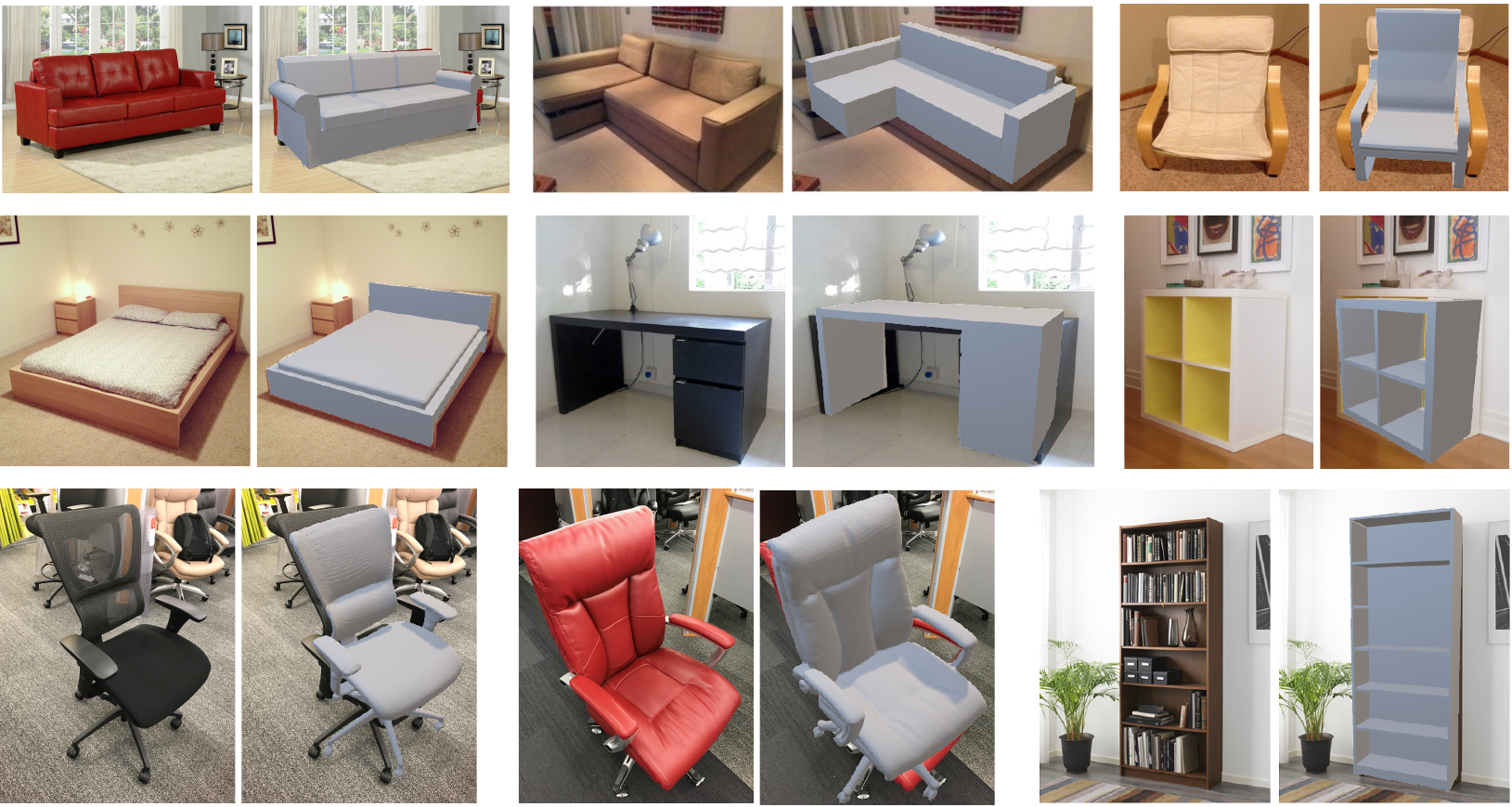}
\centering
\caption{Qualitative results on Pix3D. We visualize pose predictions by aligning 3D shapes with images.}
\label{fig:qualitative_results}
\end{figure*}

The experimental results are shown in Table \ref{fully-supervised}. We report the state-of-the-art performance of our student model compared with other image-based methods. Especially for ObjectNet3D, our method outperforms others by a significant margin (up to +5\% improvement). It suggests that keypoint-based methods \cite{grabner20183d,zhou2018starmap} may fail to capture shape information for accurate 2D-3D correspondence prediction, while contrastive learning methods \cite{Xiao2020PoseContrast} solely based on images highly depend on the pre-training effect on large external datasets like ImageNet \cite{deng2009imagenet}. In contrast, the way we gain the image representation ability by learning from 3D shapes is more effective. Moreover, although 3D shapes are coarsely aligned in Pascal3D+, we also achieve the best even close to multi-modal methods \cite{Xiao2019PoseFS}.

\noindent\textbf{Cross-Dataset Evaluation.} In order to demonstrate the generalization ability of our method, we conduct experiments in cross-dataset fashion \cite{dani20213dposelite} of category-agnostic pose estimation. We train on the 12 categories of Pascal3D+ and test on the 9 categories of Pix3D, where only 3 categories coincide with Pascal3D+. Therefore, there are 6 novel categories that are totally unseen during training. 

As shown in Table \ref{cross-dataset}, we achieve the best average performance. In particular, we even outperform the multi-modal methods \cite{dani20213dposelite,Xiao2019PoseFS}, in which \cite{Xiao2019PoseFS} adopts 3D shapes as multiple rendered views. It suggests that our method is capable of extending the implicit 3D information to enhance the performance even on novel object categories. Especially for "bookcase" and "bed", our method gains great improvement; we speculate that it is due to the simple-to-capture 3D characteristics of these two categories that maximize the correspondence with the learned 3D knowledge. 
We present qualitative results in Figure \ref{fig:qualitative_results}, where we rotate 3D shapes according to pose predictions and then align them to images.

\begin{table}[ht]
\caption{Zero-Shot and Few-Shot experimental results of category-agnostic object pose estimation on ObjectNet3D. We report results on the 20 novel categories of ObjectNet3D as defined in \cite{zhou2018starmap,tseng2019few}. All methods are based on ResNet-18.}
\begin{tabular}{lccc}
\toprule[1pt]
Method            & Setting   & Acc30 ↑       & MedErr ↓      \\ \midrule[1pt]
StarMap \cite{zhou2018starmap}          & no-shot & 0.44          & 55.8          \\ 
PoseFromShape \cite{Xiao2019PoseFS}    & no-shot & 0.55          & 41.8          \\ 
PoseContrast \cite{Xiao2020PoseContrast}     & no-shot & 0.55          & 42.7          \\ 
3DAug-Pose (Ours) & no-shot & \textbf{0.56} & \textbf{40.8} \\ \specialrule{0em}{0.5pt}{0pt} \hline \specialrule{0em}{0.5pt}{0pt}
MetaView \cite{tseng2019few}         & 10-shot   & 0.48          & 43.4          \\ 
PoseFromShape \cite{Xiao2019PoseFS}    & 10-shot   & 0.57          & 41.3          \\ 
PoseContrast \cite{Xiao2020PoseContrast}     & 10-shot   & 0.57          & 39.5          \\ 
3DAug-Pose (Ours) & 10-shot   & \textbf{0.59} & \textbf{38.9} \\ \bottomrule[1pt]
\end{tabular}
\label{zeroshot}
\end{table}

\noindent\textbf{Zero-Shot and Few-Shot Settings.} To show the robustness of our method, we enrich our evaluation in zero-shot and few-shot settings. We conduct experiments on ObjectNet3D where the 100 categories are split into 80 seen and 20 unseen categories \cite{zhou2018starmap}.



For the zero-shot experiment, following \cite{zhou2018starmap}, we train on the 80 seen categories and test on the 20 unseen categories. For the 10-shot experiment \cite{xiao2020few,tseng2019few}, the networks are first trained on the 80 seen categories, and then fine-tuned with a few labeled images from the 20 novel categories. As shown in Table \ref{zeroshot}, we achieve the best performance compared with other image-based methods. Other image-based methods aim to exploit 2D geometric similarities based on RGB images, so they heavily rely on large image training data. In contrast, our method implicitly utilize 3D shape information; such a prior knowledge of the 3D geometry is generalized on different categories, even on novel categories.

\subsection{Ablation Study}
\label{para:ablation_study}
To validate the effectiveness of our components, we conduct three sets of ablation study on ObjectNet3D.

\noindent\textbf{Ablation Study of the Teacher and the Student Baseline.} As shown in Table \ref{ablation_study}, the first 3 rows are results of our method, the teacher baseline and the student baseline. Following \cite{Xiao2019PoseFS}, the teacher baseline takes 3D shapes (point cloud) as additional input, while the student baseline only relies on images without any guidance of the teacher. In contrast, our method improves the student baseline with the Acc30 increasing from 65\% to 70\%.

\noindent\textbf{Ablation Study of Loss functions and Data augmentation.} As shown in Table \ref{ablation_study}, the ablation study results of loss functions and data augmentation are reported. We observe that all the components contribute to the model performance. The intermediate representation and the output distribution distillation are important to the performance, leading to a drop of 2\% and 3\% in Acc30 respectively when removed from 3DAug-Pose. Besides, turning off pose-related data augmentation leads to 1\% decrease in Acc30.

\noindent\textbf{Ablation Study of student image encoder Backbone.} The last two rows of Table \ref{ablation_study} present results of using different backbones as the student's image encoder. We first replace ResNet-18 in 3DAug-Pose with ResNet-50, leading to an improvement with 0.6 decrease in MedErr. It makes sense due to the larger parameter quantity of ResNet-50. We then replace ResNet-18 with VGG-11 \cite{simonyan2014very}, which has a different architecture from the proposed Contrastive Learner based on the teacher image encoder (ResNet-50). It causes 2\% decrease in Acc30 because the intermediate representation distillation through the bridge layer tends to perform worse than using similar architectures of image encoders.

\begin{table}[tp]
\caption{Ablation study on ObjectNet3D.}
\begin{tabular}{lccc}
\toprule[1pt]
Configuration        & Backbone  & Acc30 ↑ & MedErr ↓ \\ \midrule[1pt]
3DAug-Pose       & ResNet-18 & 0.70    & 25.6     \\
teacher baseline & ResNet-50 & 0.76    & 16.0     \\
student baseline & ResNet-18 & 0.65    & 31.6     \\ \specialrule{0em}{0.5pt}{0pt} \hline \specialrule{0em}{0.5pt}{0pt}
$-$ $\mathcal{L}_{CL}$+$\mathcal{L}_{KL}$        & ResNet-18 & 0.68    & 27.2     \\
$-$ $\mathcal{L}_{KD}$              & ResNet-18 & 0.67    & 31.0       \\
$-$ data augmentation  & ResNet-18 & 0.69    & 27.0       \\ \specialrule{0em}{0.5pt}{0pt} \hline \specialrule{0em}{0.5pt}{0pt}
replace ResNet-18    & ResNet-50 & 0.70    & 25.0     \\
replace ResNet-18    & VGG-11    & 0.68    & 26.7     \\ \bottomrule[1pt]
\end{tabular}
\label{ablation_study}
\end{table}

\subsection{Discussion}
\label{discussion}
\begin{figure}[htp]
\vspace{-0.5em}
\includegraphics[width=8cm]{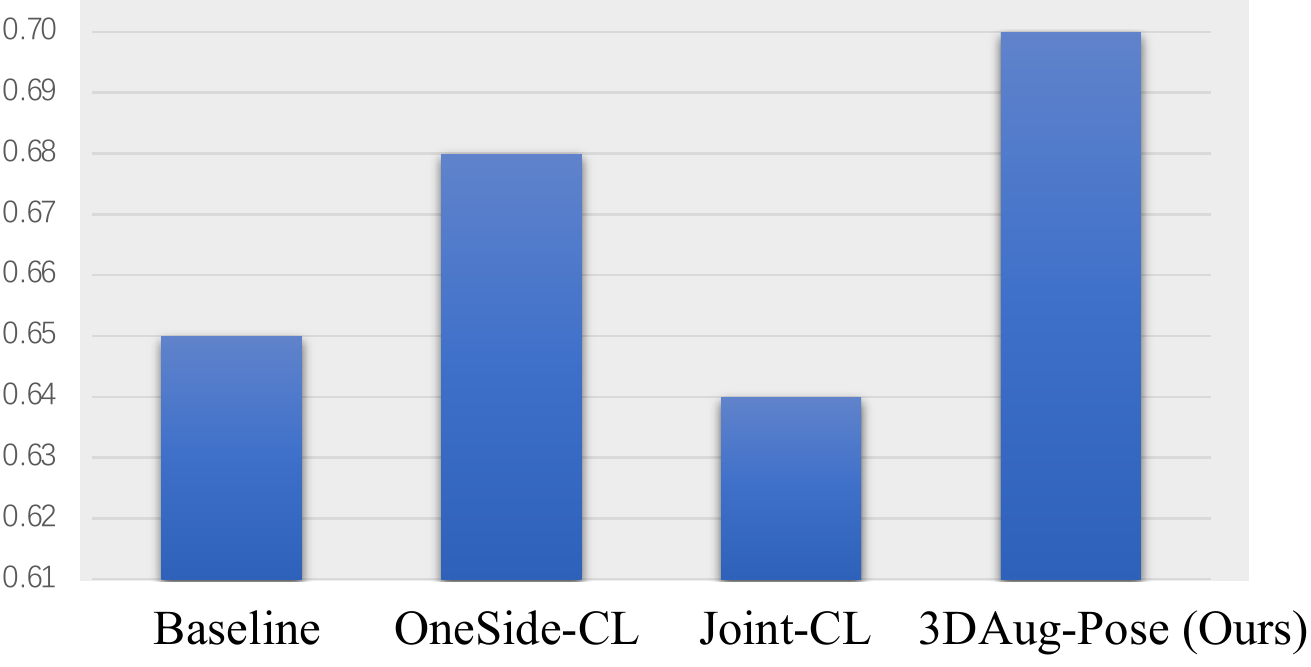}
\centering
\caption{Experimental results of methodology variants on ObjectNet3D. Results are given in Acc30.}
\label{fig:discussion}
\vspace{-1.5em}
\end{figure}

In this paper, we formulate an advanced combination strategy to integrate contrastive learning into knowledge distillation, which alleviates the withstanding gap of training procedures between these two approaches. In order to observe and compare effects of different combination strategies, we investigate two approaches as methodology variants:

\begin{itemize}[leftmargin=*]
\item \textbf{OneSide-CL:} this is the most common strategy \cite{tian2019contrastive, sun2020contrastive, dai2021learning} that fuses contrastive learning into the second training stage. Specifically, we first train the teacher model for pose estimation, then we freeze the pre-trained teacher and transfer its knowledge, where we conduct one-side contrastive learning that back-propagates only in the student model;
\item \textbf{Joint-CL:} this is the most straightforward strategy that conducts contrastive learning on the teacher and the student jointly at the first training stage. Specifically, we combine contrastive learning and pose estimation as a joint objective, training the teacher to transfer its representation knowledge to the student while optimizing for the pose estimation task. At the second training stage, we fine-tune the pre-trained student for pose estimation with the output guidance of the teacher.
\end{itemize}
The results of different combination strategies are reported in Figure \ref{fig:discussion}, in which Baseline refers to the image-based student model without any guidance of the teacher. As the result shows, OneSide-CL exhibits inferior performance compared with the proposed 3DAug-Pose. We find it more difficult for the student to obtain cross-modal features from the teacher without back-propagating in both models jointly. For Joint-CL, it even drops 1\% in Acc30 compared with Baseline. We find that involving the student for joint contrastive learning affects the downstream performance of the teacher (2\% decrease in Acc30), which tends to perform worse in the further knowledge distillation. In contrast, introducing our Contrastive Learner grounded on the teacher model, we effectively conduct joint contrastive learning while preserving great representation ability of the teacher. Thereby, at the second training stage, we best distill cross-modal knowledge via knowledge distillation.

\section{CONCLUSION}


In this paper, we take advantage of existing 3D shapes of training data to enhance the performance on RGB images for category-agnostic object pose estimation. Technically, we propose a novel contrastive knowledge distillation framework to transfer 3D knowledge learned by a multi-modal model to an image-based model. Moreover, we leverage a Contrastive Learner as the ``bridge'' for effectively shifting 3D-augmented image representations across two models. Our framework provides a solution to best distill privileged cross-modal information in intermediate representation and output distribution. Extensive experiments on ObjectNet3D, Pascal3D+ and Pix3D show the effectiveness of our method.

\section{ACKNOWLEDGEMENT}
This work was supported by the National Key Research and Development Program of China, No.2018YFB1402600.

\bibliographystyle{IEEEtran}
\bibliography{MyPaper.bib}

\end{document}